\documentclass[letterpaper, 10 pt, conference]{ieeeconf}  

\IEEEoverridecommandlockouts                              

\title{\LARGE \bf
STEAM: A Training-Free Congestion-Aware Enhancement Framework for Decentralized  Multi-Agent Path Finding}

\author{Mingyang Feng, Mengnuo Zhang, Shaoyuan Li and Xiang Yin
\thanks{This work was supported by the National Natural Science Foundation of China (62173226, 92367203). 
 }
\thanks{M. Feng, M. Zhang, S. Li and X. Yin  are  with the School of Automation and Intelligent Sensing, Shanghai Jiao Tong University, Shanghai 200240, China. (Corresponding Author: Xiang Yin) {\tt  E-mail: \{Fmy-135214,130638200702058512,syli,yinxiang\}\newline @sjtu.edu.cn}.}  
}

\usepackage[algo2e,ruled,linesnumbered,vlined,boxed,commentsnumbered]{algorithm2e}
\usepackage{amsmath}
\usepackage{multirow}
\usepackage{array}
\usepackage{graphicx}
\usepackage{cite}
\let\labelindent\relax
\usepackage{enumitem}
\usepackage{algorithmic}
\usepackage{hyperref}
\usepackage{caption}
\usepackage{float} 
\usepackage{subcaption}
\usepackage{threeparttable}
\usepackage{bm}
\usepackage{amssymb}
\usepackage{xcolor}
\usepackage{booktabs}
\usepackage{multirow}

\usepackage{flushend}

\begin{document}

\maketitle
\thispagestyle{empty}
\pagestyle{empty}

\begin{abstract}
We propose \textbf{STEAM} (\underline{\textbf{S}}patial, \underline{\textbf{T}}emporal, and 
\underline{\textbf{E}}mergent congestion \underline{\textbf{A}}wareness for \underline{\textbf{M}}APF), a training-free
test-time enhancement framework for learning-based decentralized Multi-Agent Path Finding
(MAPF) in discrete environments. Given a pretrained decentralized policy, STEAM requires
no retraining, architectural modification, or replacement by a centralized planner. Instead,
it injects lightweight congestion-aware guidance into the original policy execution. STEAM
first rolls out the shortest paths induced by the current cost-to-go maps to identify
potential future congestion hotspots. Spatially avoidable congestion is mitigated by
updating agent-specific cost-to-go information, while spatially unavoidable bottlenecks are
handled through temporal logit correction. In addition, emergent local congestion is
reduced by a density-aware logit correction based on neighboring agents' corrected
cost-to-go maps. Extensive experiments on representative learning-based decentralized MAPF
algorithms show that STEAM consistently improves success rate, makespan, and solution cost,
with success-rate gains of up to 60\% and only minor computational overhead. The
implementation is available at
\url{https://anonymous.4open.science/r/STEAM-MAPF-7A62}. 
\end{abstract}

\section{Introduction}
Multi-Agent Path Finding (MAPF) aims to compute collision-free paths for multiple agents that move in a shared environment from their initial locations to their individual goals. It is a fundamental problem in multi-robot coordination and has broad applications in warehouse automation~\cite{stern2019multi}, robotic swarm systems~\cite{agaskar2025deepfleet}, and game AI~\cite{wang2025intelligent}. Despite its simple formulation, MAPF is computationally challenging. Finding optimal solutions is NP-hard \cite{stern2019multi,fioravantes2025solving}, and the difficulty becomes more severe in large-scale or highly congested scenarios where many agents interact in narrow passages, intersections, and other shared regions.

Classical MAPF methods are mainly based on centralized global search. Representative algorithms, such as Conflict-Based Search (CBS) and its variants, explicitly reason about conflicts among agents and can provide strong completeness or optimality guarantees \cite{li2019multi}. Other approaches, such as prioritized planning \cite{erdmann1987multiple,silver2005cooperative}, improve computational efficiency by planning agents sequentially, but often sacrifice completeness and robustness. More recent methods, including EECBS~\cite{li2021eecbs} and LaCAM~\cite{okumura2023lacam}, further improve the trade-off between solution quality and planning efficiency. Nevertheless, these centralized approaches still face intrinsic scalability challenges, because the underlying planning problem is coupled through the joint state space of all agents, whose size grows combinatorially with the number of agents. This makes it difficult to directly apply centralized global search to very large multi-agent systems in real time.
 
To address the computational burden of centralized planning, decentralized MAPF methods have received increasing attention in recent years, especially those leveraging machine learning techniques. In these methods, each agent is treated as an independent decision maker that selects actions based on its local observation and limited information about nearby agents, such as their positions, intents, or cost-to-go maps. A shared policy, typically implemented by a deep neural network, is trained to map such local information to action decisions or action logits. Representative learning-based decentralized MAPF methods include PRIMAL/PRIMAL2~\cite{sartoretti2019primal,damani2021primal}, MAGAT~\cite{li2021message}, and MAPF-GPT~\cite{andreychuk2025mapf,andreychuk2025advancing}. Recent studies also suggest that the learned policies often behave like local shortest-path followers that move agents toward their goals while avoiding obstacles and nearby agents \cite{lahire2026interpretable}. Compared with centralized methods, these decentralized policies are highly scalable, since the same policy can be executed independently by all agents and does not require explicit search over the full joint state space.

However, purely decentralized execution has an inherent limitation: each agent only observes local information and communicates only with nearby agents, lacking global awareness of future congestion. Agents may independently move toward the same bottleneck region before becoming aware of each other, causing severe congestion or deadlock. Such situations cannot be reliably prevented using purely local observations alone. Therefore, although local coordination is essential for scalability, it remains insufficient for anticipating global congestion patterns. Some prior methods address this issue reactively by switching from the learned decentralized policy to a conventional global planner, such as "LaCAM", once congestion or deadlock is detected \cite{wang2025deadlock}. This strategy resolves congestion only after it has already formed, introducing extra waiting, frequent replanning, and additional solver overhead, while partially weakening the scalability benefits of decentralized execution.

\begin{figure*}[t]
    \centering
    \includegraphics[width=1\linewidth]{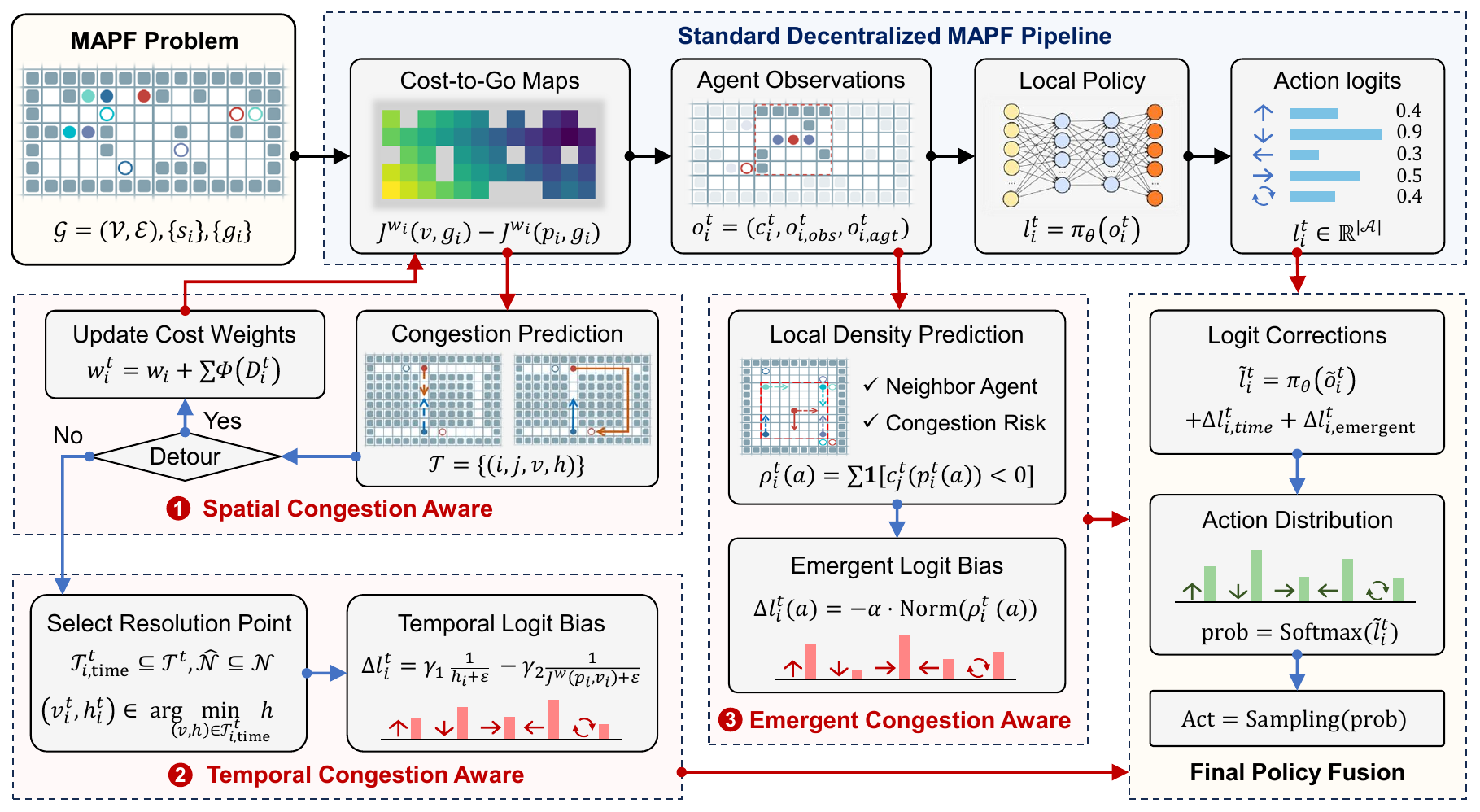}
    \caption{Overview of STEAM. The \textbf{S}patial-, \textbf{T}emporal-, and \textbf{E}mergent congestion-\textbf{A}ware modules for \textbf{M}APF capture congestion from global and local perspectives. Black arrows denote the original decentralized MAPF policy pipeline. Red arrows indicate interactions between the proposed modules and the original pipeline, while blue arrows denote our extensions.}
    \label{fig:model}\vspace{-6pt}
\end{figure*}
In this paper, we propose a training-free and policy-agnostic congestion-aware enhancement framework for learning-based decentralized MAPF. Given any pretrained third-party decentralized policy, our method does not modify its network architecture, retrain its parameters, or replace it with a centralized planner. Instead, it provides lightweight global and local guidance during test-time execution. Specifically, we first roll out the shortest paths induced by the current cost-to-go maps and use them to identify potential future congestion points. For congestion that can be spatially avoided, we modify the corresponding agent-specific cost-to-go map, thereby changing the local observation provided to the decentralized policy and guiding the agent to bypass the congested region. For congestion that cannot be resolved spatially, typically in bottleneck regions, we apply a temporal logit correction to encourage one agent to delay or take a less aggressive action. Finally, we incorporate a local density-based correction using neighboring agents' corrected cost-to-go maps to discourage movements toward crowded regions. In this way, our method injects global congestion awareness and local crowding awareness into the original decentralized execution process while preserving the learned policy itself.

The proposed framework has several important advantages. First, it is training-free and plug-and-play: it can be integrated with existing learning-based decentralized MAPF policies without additional data collection, retraining, or architecture modification. Second, it is computationally lightweight, since it only rolls out individual shortest paths and applies sparse cost-map and logit corrections, rather than searching over the full joint state space of all agents. Third, it is policy-agnostic and generalizes across different learned  policies, as long as the policy takes local observations with cost-to-go information and outputs action preferences. We validate these properties on representative decentralized MAPF baselines, including MAPF-GPT and PRIMAL2. Experimental results show that our method consistently improves success rate and coordination efficiency, especially in dense and congestion-prone environments, while introducing only minor computational overhead.

\section{Problem Formulation}

\subsection{Multi-Agent Path Finding}

We consider a standard Multi-Agent Path Finding (MAPF) problem on an undirected graph
$\mathcal{G}=(\mathcal{V},\mathcal{E})$, where $\mathcal{V}$ is the vertex set and
$\mathcal{E}\subseteq \mathcal{V}\times\mathcal{V}$ is the edge set. In discrete MAPF scenarios, the environment is modeled as a grid map, from which the graph $\mathcal{G}$ is induced. Edges corresponding to
blocked movements caused by static obstacles are excluded from $\mathcal{E}$. Let
$\mathcal{N}=\{1,\ldots,N\}$ denote the set of agents. Each agent $i\in\mathcal{N}$ has a start vertex $s_i\in\mathcal{V}$ and a goal vertex $g_i\in\mathcal{V}$. 
A path of agent $i$
is denoted by $P_i=(p_i^0,p_i^1,\ldots)$, where $p_i^t$ is the position of agent $i$ at
time step $t$.

At each time step, an agent can either move to an adjacent vertex or wait at its current
vertex, i.e.,
\[
p_i^{t+1}\in \{p_i^t\}\cup\{v\in\mathcal{V}\mid (p_i^t,v)\in\mathcal{E}\}.
\]
A set of paths $P=\{P_1,\ldots, P_N\}$ is feasible if all agents reach their goals and no
collisions occur. Specifically, for any agent $i\!\neq \!j$ and   time step $t$, feasible paths must
satisfy
\[
p_i^t\neq p_j^t\quad \text{ and }\quad 
(p_i^t,p_i^{t+1})\neq(p_j^{t+1},p_j^t),
\]
which exclude collisions and edge-swap conflicts, respectively.
The objective is to find feasible paths that minimize a given performance criterion, such
as   sum of costs. 

\subsection{Learning-Based Decentralized MAPF}

Decentralized MAPF methods avoid explicit joint-state search by letting agents make decisions independently based on local observations. For homogeneous agents, a common practice is to use a \emph{shared policy} across all agents.
Let $p_i^t$ denote the current
position of agent $i$ at time step $t$, and let $\mathcal{W}_i^t \subseteq \mathcal{V}$ denote
the set of vertices contained in the local observation window of agent $i$ at time $t$. In our setting, $\mathcal{W}_i^t$ corresponds to an $R \times R$ grid centered at
$p_i^t$.
The local observation is typically represented as a multi-channel tuple. Although the exact
channel design may vary across different decentralized MAPF models, it usually contains
local obstacle information, neighboring-agent information, and a cost-to-go channel. We
write it abstractly as
\[
o_i^t =
\Big(
o_{i,\mathrm{obs}}^t,\,
o_{i,\mathrm{agt}}^t,\,
C_i^t
\Big).
\]
More specifically, for each agent $i$, let
$w_i:\mathcal{V}\rightarrow \mathbb{R}_{>0}\cup\{\infty\}$ be an
agent-specific vertex weight function, where $w_i(v)$ denotes the
traversal cost of vertex $v$. For the vertices corresponding to static
obstacles, we define $w_i(v)=\infty$, indicating that these vertices
are not traversable. Given the goal vertex $g_i$, the global
cost-to-go value of vertex $v\in\mathcal{V}$ under $w_i$ is denoted by
$J^{w_i}(v,g_i)$ and is defined as the weighted shortest-path cost from
$v$ to $g_i$.
The local cost-to-go channel used in the observation is then the restriction of this global
map to the observation window:
\[
C_i^t(v)=J^{w_i}(v,g_i) - J^{w_i}(p_i^t,g_i), \qquad v\in\mathcal{W}_i^t .
\]
In the standard unweighted MAPF setting, one can simply set $w_i(v)=1$ for every free
vertex $v$, in which case $J^{w_i}(v,g_i)$ reduces to the shortest-path distance from $v$
to $g_i$.

In learning-based MAPF, each agent is controlled by a pretrained decentralized policy,
typically implemented as a deep neural network. At each time step, the policy maps the
local observation to \emph{action logits}:
\[
\ell_i^t=\pi_\theta(o_i^t)\in\mathbb{R}^{|\mathcal{A}|},
\]
where $\mathcal{A}$ is the discrete action set and $\ell_i^t$ encodes the preferences over
candidate actions. The final action is selected from the action distribution obtained
by applying a Softmax function to $\ell_i^t$, either deterministically or through stochastic sampling during execution.

In conventional decentralized execution, the weight function $w_i$ is fixed before deployment and does not depend on the other agents. Therefore, the global cost-to-go map  only needs to be computed once for each agent and can be reused throughout
execution. This design is scalable, since each agent only queries the local restriction of its precomputed cost-to-go map. However, because the fixed map ignores the future motion of
other agents, locally reasonable decisions may collectively induce congestion in narrow passages, intersections, or other high-density regions.

This motivates the  policy-agnostic enhancement problem. Given an existing
third-party decentralized MAPF policy $\pi_\theta$, we aim to develop a lightweight
test-time mechanism that improves deployment-time performance while preserving the
original policy. 
The policy parameters and network architecture  remain unchanged.
Instead, we only modify the test-time information
provided to the policy, such as the cost-to-go channel induced by $w_i$, and the final action logits. 
Therefore, different learning-based decentralized
MAPF solvers can be used as fixed backbone decision modules without model-specific
retraining or fine-tuning.

  \section{Methodology}

This section presents the proposed training-free congestion-aware enhancement framework
for learning-based decentralized MAPF. The framework keeps the pretrained policy
$\pi_\theta$ unchanged and only modifies its test-time execution. The key idea is to
resolve potential congestion at three complementary levels.
\begin{itemize}
    \item 
    First, if a predicted congestion spot can be avoided spatially, we guide one of the
involved agents to bypass that spot by modifying its agent-specific map weight $w_i$.
This changes the induced cost-to-go map $J^{w_i}$ and hence the local cost-to-go channel in the observation $o_i^t$. 
    \item 
    Second, if the congestion cannot be avoided spatially,
which usually indicates that the involved agents share a narrow mandatory passage, we
resolve it temporally by modifying the action logits and encouraging one agent to delay or choose an alternative immediate action. 
    \item 
    Third, for agents within the local observation
range, we further use the corrected cost-to-go maps of neighboring agents to estimate
locally crowded regions and apply an additional logit correction to avoid emergent
high-density movements.
\end{itemize}

Formally, at time step $t$, the original policy produces $\ell_i^t=\pi_\theta(o_i^t)$ and our method constructs a corrected observation $\tilde{o}_i^t$ by updating the cost-to-go
channel when spatial congestion resolution is possible, and then applies two logit-level
corrections:
\[
\tilde{\ell}_i^t
=
\pi_\theta(\tilde{o}_i^t)
+
\Delta \ell_{i,\mathrm{time}}^t
+
\Delta \ell_{i,\mathrm{emergent}}^t .
\]
Here, $\Delta \ell_{i,\mathrm{time}}^t$ is induced by globally predicted temporal
congestion, and $\Delta \ell_{i,\mathrm{emergent}}^t$ is induced by locally estimated
density.

\subsection{Global Spatial Congestion Resolution}

We first predict potential future congestion by rolling out the shortest paths induced by the precomputed cost-to-go maps $\{J^{w_i}\}_{i\in \mathcal{N}}$. Importantly, this procedure does not detect congestion that has already occurred; instead, it anticipates possible future conflicts, assuming that each agent continues to follow its current shortest path toward the goal. 

At time step $t$, for each agent $i$, we compute a shortest path from its
current position $p_i^t$ to its goal $g_i$ under the static weight function $w_i$. This
path is denoted by $P_i^\star$. Let $P_i^\star(h)$ denote the vertex reached by agent $i$
after $h$ steps along this path. If the path has already reached the goal, we assume that
the agent remains at the goal.

Based on these predicted paths, we define the set of spatiotemporal congestion points as
\[
\mathcal{T}^t
=
\left\{
(i,j,v,h)
\;\middle|\;
i<j,\;
P_i^\star(h)=P_j^\star(h)=v
\right\}.
\]
Each element $(i,j,v,h)\in\mathcal{T}^t$ indicates that agents $i$ and $j$ are predicted
to occupy the same vertex $v$ at future time offset $h$. These points are potential future
congestion spots.

For each congestion point $(i,j,v,h)\in\mathcal{T}^t$, we test whether it can be avoided
spatially. Specifically, for each involved agent $r\in\{i,j\}$, we construct a temporary
probing weight function
\[
\hat{w}_{r}(u)
=
w_r(u)
+
\Lambda\mathbf{1}_{\{u=v\}},
\]
where $\Lambda>0$ is a large penalty. The temporary weight
$\hat{w}_{r}$ is used only for this congestion point and is reset to  $w_r$ when testing the next congestion point. We then recompute the shortest path
from $p_r^t$ to $g_r$ under $\hat{w}_{r}$, denoted by $\hat{P}_{r}^{\star}$.

If neither agent changes its shortest path, i.e.,
\[
\hat{P}_{i}^{\star}=P_i^\star
\quad \text{and} \quad
\hat{P}_{j}^{\star}=P_j^\star,
\]
then this means that the congestion cannot be avoided by a spatial detour. In this case, we add the
congestion point $(i,j,v,h)$ to the \emph{temporal-congestion set}
$\mathcal{T}_{\mathrm{time}}^t\subseteq\mathcal{T}^t$, which will be handled latter.
Otherwise, at least one involved agent can bypass the congested vertex. For each agent
$r\in\{i,j\}$ whose path changes, we compute the detour cost as
\[
D_r^t(v)
=
J^{\hat{w}_{r}}(p_r^t,g_r)
-
J^{w_r}(p_r^t,g_r).
\]
We then select the agent with the smaller detour cost:
\[
k
=
\arg\min_{r\in\{i,j\}:\hat{P}_{r}^{\star}\neq P_r^\star}
D_r^t(v).
\]
The selected pair $(v,D_k^t(v))$ is stored as a spatial intervention for agent $k$.

After all congestions have been examined, we aggregate all selected spatial
interventions $\{(v,D_i^t(v))\}$. The updated weight function
for agent $i$ at time step $t$ is defined as
\[
w_i^t(u)
=
w_i(u)
+
\sum_{(v,D_i^t(v))\in\mathcal{S}_i^t}
\phi(D_i^t(v))\mathbf{1}_{\{u=v\}} .
\]
Here, $\phi(\cdot)$ is an increasing penalty function. In our implementation, we use the
quadratic form, which assigns stronger penalties to more costly congestion spots and helps discourage
repeated oscillatory rerouting.

Finally, each agent recomputes its local cost-to-go channel based on the updated weight
function $w_i^t$. Specifically, the corrected cost-to-go channel is obtained by 
$\tilde{C}_i^t(u)=J^{w_i^t}(u,g_i)-J^{w_i^t}(p_i^t,g_i)$ for all $u\in\mathcal{W}_i^t$ in the current observation window. 
The corrected local observation $\tilde{o}_i^t$ is obtained by replacing the original cost-to-go channel $C_i^t$ with $\tilde{C}_i^t$. Therefore, spatial congestion is resolved
by modifying the input information provided to the pretrained policy, while the policy
parameters remain unchanged.

Importantly, modifying the observation representation in general often changes the underlying input distribution, which may require redesigning the model architecture or retraining the policy to adapt to the new observations. In contrast, our approach preserves the original structure and semantics of the cost-to-go representation. The modified channel $\tilde{C}_i^t$ still provides a valid cost-to-go signal toward the goal, but incorporates additional congestion-aware spatial information through the adaptive weight function $w_i^t$. As a result, the policy continues to receive observations that remain consistent with its original training distribution and decision-making mechanism, avoiding severe out-of-distribution effects. Conceptually, the guidance induced by the cost-to-go channel is shifted from shortest-path-oriented navigation toward congestion-aware suboptimal routing, while preserving the original behavioral prior learned by the pretrained policy. Therefore, the pretrained policy can be directly reused without any architectural modification or additional training.

\subsection{Global Temporal Congestion Resolution}

The spatial module separates the detected congestions into two classes. If a
congestion point can be avoided by modifying the map weight of one involved agent, it is
handled by the spatial update in the previous subsection. Otherwise, the congestion point
is added to the residual temporal-congestion set
$\mathcal{T}_{\mathrm{time}}^t$. For any
$(i,j,v,h)\in\mathcal{T}_{\mathrm{time}}^t$, penalizing the congested vertex $v$ does not
change the shortest path of either involved agent. This indicates that the agents are
likely to share a mandatory bottleneck, such as a narrow passage or a one-cell corridor.
In such cases, creating an artificial spatial detour in the cost-to-go map is ineffective. Moreover, since no feasible suboptimal spatial route exists around the bottleneck, forcing additional cost perturbations may distort the original cost-to-go structure and produce out-of-distribution observations that are inconsistent with the policy's training regime.
We therefore resolve these residual congestions by temporally separating the involved
agents through logit-level correction.

We first select the agents to be temporally adjusted.  We  select a
subset of agents $\hat{\mathcal{N}}^t\subseteq\mathcal{N}$ such that every unresolved congestion pair is covered, i.e., 
\[
\forall (i,j,v,h)\in\mathcal{T}_{\mathrm{time}}^t: 
\{i,j\}\cap\hat{\mathcal{N}}^t\neq\emptyset .
\]
Among all such subsets, we choose one with the smallest cardinality, so that the temporal
intervention affects as few agents as possible. This selection can be obtained by a small integer program.

For each selected agent $i\in\hat{\mathcal{N}}^t$, we focus on its earliest unresolved
temporal congestion. Let
\[
\mathcal{T}_{i,\mathrm{time}}^t
\!=\!
\left\{
(v,h)
\;\middle|\;
\exists j\! :\! (i,j,v,h)\!\in\!\mathcal{T}_{\mathrm{time}}^t
\vee  (j,i,v,h)\!\in\!\mathcal{T}_{\mathrm{time}}^t
\right\}.
\]
We choose
\[
(v_i^t,h_i^t)
\in
\arg\min_{(v,h)\in\mathcal{T}_{i,\mathrm{time}}^t} h .
\]
The reason for only considering the earliest unresolved congestion is that the proposed
method is executed online in a receding-horizon manner. Once an earlier bottleneck
conflict is delayed or removed, later predicted conflicts involving the same agent may
also disappear after the next prediction and update step.

We then modify the logits of the selected agents to discourage actions that move them
toward the unresolved bottleneck too aggressively. Let
$p_i^t(a)=\mathrm{Next}(p_i^t,a)$ 
denote the next vertex reached by agent $i$ at position $p_i^t$ if action $a\in\mathcal{A}$ is executed.   For the selected congestion vertex $v_i^t$, define
the one-step progress of action $a$ toward $v_i^t$ as
\[
\Delta_i^t(a)
=
J^{w_i}(p_i^t,v_i^t)
-
J^{w_i}(p_i^t(a),v_i^t).
\]
A positive value of $\Delta_i^t(a)$ means that action $a$ moves agent $i$ closer to the
unresolved bottleneck, while a non-positive value means that the action either waits  or moves away from the bottleneck.

The temporal logit correction is defined as
\[
\Delta \ell_{i,\mathrm{time}}^t(a)
=
-
\mathbf{1}_{\{i\in\hat{\mathcal{N}}^t\}}
\lambda_i^t
\left[\Delta_i^t(a)\right]_+ ,
\]
where $[x]_+=\max\{x,0\}$ and
\[
\lambda_i^t
=
\gamma_{\mathrm{time}}\frac{1}{h_i^t+\varepsilon}
+
\gamma_{\mathrm{dist}}\frac{1}{J^{w_i}(p_i^t,v_i^t)+\varepsilon}.
\]
Here, $\gamma_{\mathrm{time}}>0$ and $\gamma_{\mathrm{dist}}>0$ are tunable coefficients,
and $\varepsilon>0$ is a small constant used to avoid division by zero. This correction
penalizes actions that make positive progress toward the unresolved congestion vertex. The
penalty becomes stronger when the congestion is closer in time or closer in space, since
such congestion is more reliable and more urgent. In contrast, waiting actions or actions
that do not move toward the bottleneck are not penalized, thereby encouraging a temporal
offset between the involved agents.

The logits after temporal correction are therefore
\[
\ell_{i,\mathrm{time}}^t(a)
=
\pi_\theta(\tilde{o}_i^t)(a)
+
\Delta \ell_{i,\mathrm{time}}^t(a),
\qquad a\in\mathcal{A}.
\]
This temporal module complements the spatial module: spatially resolvable congestion is
handled by updating the cost-to-go channel in the observation, whereas spatially
unresolvable congestion is handled by directly reshaping the action preferences at the
current decision step. Thus, the pretrained policy remains unchanged, while its execution
is biased away from imminent bottleneck conflicts.

\subsection{Local Emergent Congestion Resolution}

Predicting global spatiotemporal congestion in advance enables system-level congestion mitigation.
However, its effect may diminish when many agents become crowded in a local region as an emergent behavior during
online execution. Our goal is to proactively detect potential local
crowding and reduce the formation of large agent aggregations before they become difficult
to resolve. To this end, we introduce a local emergent congestion-resolution module that further
adjusts the action logits using the corrected local cost-to-go maps of neighboring agents.

Let $\mathcal{N}_i^t$ denote the set of neighboring agents observed by agent $i$ in its
local observation window at time $t$. For each candidate action $a\in\mathcal{A}$, we
evaluate the local congestion risk associated with the successor vertex $p_i^t(a)$. The
key idea is to estimate whether this candidate successor vertex is also attractive to
nearby agents. Since the spatial module has produced an updated weight function $w_j^t$
for each neighboring agent $j$, we use the corresponding corrected cost-to-go map
$C_j^t(\cdot)$ to infer the local motion tendency of agent $j$.

For each neighboring agent $j\in\mathcal{N}_i^t$, we define
\[
\Delta J_{j\rightarrow i}^t(a)
=
C_j^t(p_i^t(a)).
\]
If $\Delta J_{j\rightarrow i}^t(a)<0$, then the candidate successor vertex $p_i^t(a)$
has a lower cost-to-go value for agent $j$ than its current position $p_j^t$. This
indicates that agent $j$ may also tend to move toward this region. Therefore, we define
the local density score induced by action $a$ as
\[
\hat{\rho}_i^t(a)
=
\sum_{j\in\mathcal{N}_i^t}
\mathbf{1}
\left[
\Delta J_{j\rightarrow i}^t(a)<0
\right].
\]
A larger $\hat{\rho}_i^t(a)$ means that more neighboring agents may be attracted to the
candidate successor vertex of agent $i$, and hence action $a$ is more likely to lead to a
locally contested region.

We then penalize actions with high estimated local density.
To make the local correction compatible with the scale of the policy output $\ell_{i,\mathrm{time}}^t(a)$ after global temporal correction, we compute
the standard deviation of these logits:
$\bar{\ell}_{i,\mathrm{time}}^t
=
\frac{1}{|\mathcal{A}|}
\sum_{a\in\mathcal{A}}
\ell_{i,\mathrm{time}}^t(a)
$
and $
\sigma_i^t
=
\text{sqrt}(
\frac{1}{|\mathcal{A}|}
\sum_{a\in\mathcal{A}}
\left(
\ell_{i,\mathrm{time}}^t(a)
-
\bar{\ell}_{i,\mathrm{time}}^t
\right)^2)$. 
The local logit correction is then defined as
\[
\Delta \ell_{i,\mathrm{emergent}}^t(a)
=
-
\alpha \sigma_i^t \hat{\rho}_i^t(a),
\qquad a\in\mathcal{A},
\]
where $\alpha>0$ is a tunable coefficient controlling the strength of the local
intervention. The final enhanced logits are  
\[
\tilde{\ell}_i^t(a)
=
\ell_{i,\mathrm{time}}^t(a)
+
\Delta \ell_{i,\mathrm{emergent}}^t(a),
\qquad a\in\mathcal{A}.
\]
This correction decreases the preference of actions that move agent $i$ toward regions
also attractive to many neighboring agents. Therefore, the local module mitigates emergent
short-range crowding without modifying the policy parameters or invoking an additional
global planner. Together with the global spatial and temporal modules, it provides a
lightweight test-time mechanism for congestion-aware decentralized execution.




 \subsection{Complexity Discussion}

The proposed framework is designed as a lightweight test-time enhancement for pretrained
decentralized MAPF policies. It does not perform centralized joint-state search, does not
retrain the policy, and does not invoke an additional global planner during execution. The
main extra computation comes from detecting predicted congestion points, testing whether a
small number of such points can be spatially bypassed, and applying logit-level corrections
for temporal and local congestion. Since these operations are performed on predicted paths,
detected congestion points, and local neighborhoods rather than on the full joint state
space, the additional cost remains moderate.

In practice, congestion points are usually sparse and concentrated around bottleneck
regions such as narrow passages and intersections. Spatial updates only modify the weights
of selected agents at selected vertices, and the resulting cost-to-go maps can be updated
incrementally. Temporal correction only affects a small subset of agents involved in
spatially unresolvable conflicts, while local correction only requires checking neighboring
agents within the observation window. Therefore, the overall computation grows with the
number of detected congestion points and local neighbors, rather than exponentially with
the number of agents. This makes the method suitable for online deployment as a
policy-agnostic and computationally lightweight enhancement. 

\section{Experiment}

\begin{figure*}
    \centering
    \includegraphics[width=1\linewidth]{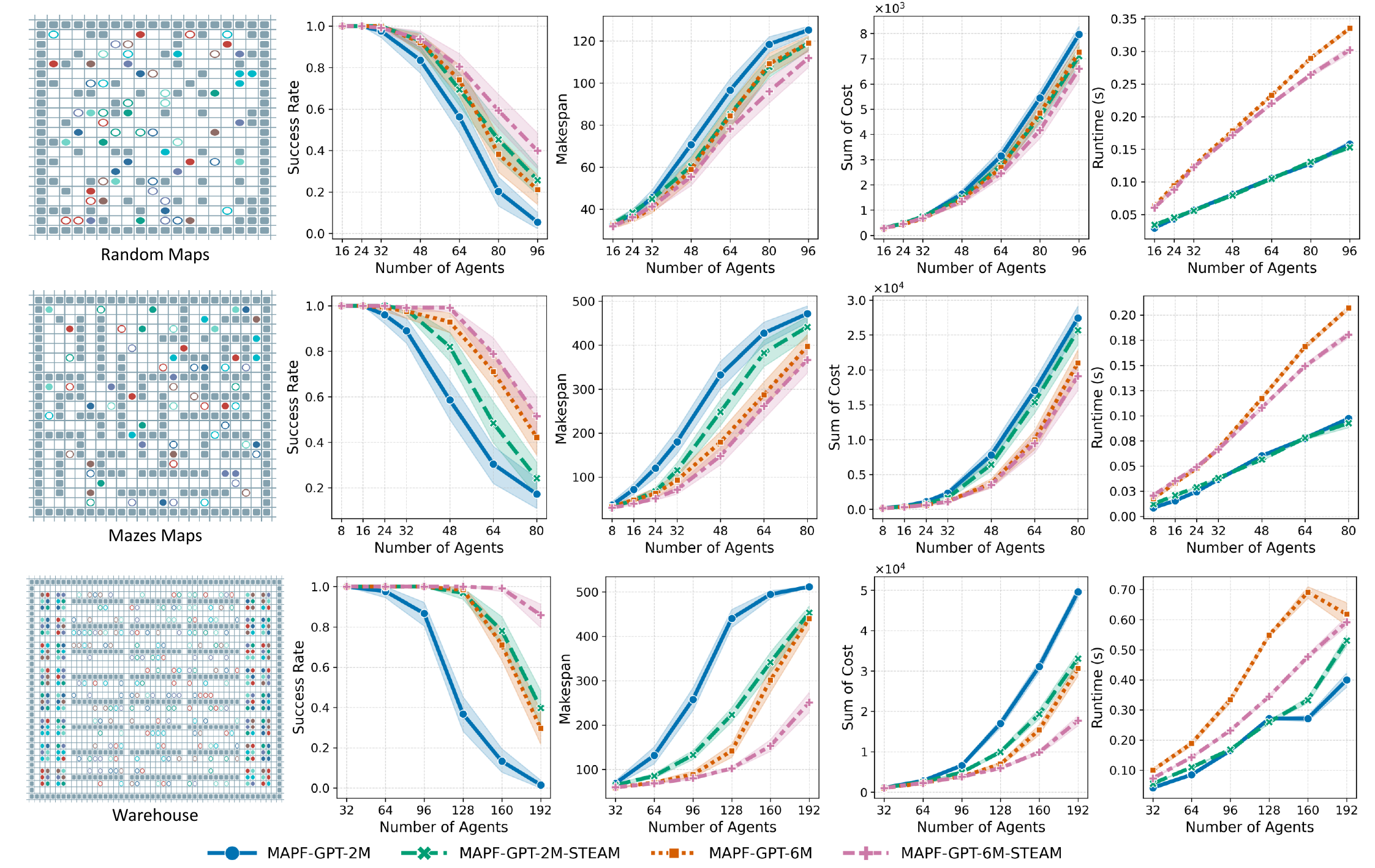}
    \caption{Optimization performance of MAPF-GPT across different scenarios. The first column shows the map type, and the following columns report the corresponding metrics on each map. Entries with the suffix ``-STEAM'' indicate results obtained after integrating our proposed framework into the baseline model.}
    \label{fig:result}
\end{figure*}

\subsection{Experimental Setup}

We evaluate the proposed framework on two representative learning-based decentralized
MAPF baselines, PRIMAL2~\cite{damani2021primal} and MAPF-GPT~\cite{andreychuk2025mapf},
to demonstrate its effectiveness. Both baselines generate
action logits from local observations that include cost-to-go information, making them
compatible with our congestion-aware observation and logit correction framework. For
MAPF-GPT, we consider two lightweight variants, MAPF-GPT-2M and MAPF-GPT-6M, which provide
different model capacities. For PRIMAL2, we use the original architecture designed for
high-density structured environments. In all cases, our method is integrated into the
original inference pipeline without retraining, fine-tuning, or modifying the policy
network architecture.

Following~\cite{andreychuk2025mapf}, we evaluate the models on three representative map
types: \textbf{random}, \textbf{maze}, and \textbf{warehouse}. 
To better expose congestion effects and
stress-test the proposed method, we use denser agent configurations than the original
settings in several scenarios.

\begin{itemize}
    \item \textbf{Random}: Agents are uniformly distributed on the map, leading to
    unstructured interactions. We increase the number of agents from 64 to 96 to create
    denser traffic.

    \item \textbf{Maze}: Agents navigate through narrow corridors and structured
    obstacles, where bottleneck congestion frequently occurs. For MAPF-GPT, we increase
    the number of agents from 64 to 80. For PRIMAL2, we additionally evaluate dense maze
    settings on $40\times40$ maps with obstacle densities ranging from 0.3 to 0.8 and
    agent counts of 32, 64, 128, 146, and 160.

    \item \textbf{Warehouse}: Agents operate in structured aisle-like environments that
    resemble real-world logistics scenarios. For MAPF-GPT, we follow the original
    warehouse setting. For PRIMAL2, we construct a larger warehouse environment supporting
    up to 192 agents and evaluate configurations with 128, 160, 172, 184, and 192 agents.
\end{itemize}

All experiments are conducted over 128 episodes with randomly sampled obstacle layouts,
start-goal configurations, and agent placements. Unless otherwise specified, the
hyperparameters of our method are set to $\alpha=0.3$, and
$\gamma_{time}=\gamma_{dist}=4$. The congestion-aware update is applied every 5 time steps. All
simulations are implemented in Pogema~\cite{skrynnik2024pogema} and executed on a machine
equipped with an AMD Ryzen 9 CPU and an NVIDIA RTX 5090 GPU. 

\subsection{Experimental Results}

\begin{table*}[t]
\centering
\caption{Performance comparison on Dense maze and Warehouse environments with PRIMAL2. ``w/o'' denotes the baseline results, while ``w/STEAM'' denotes the results after integrating our proposed algorithm. All reported values are presented as mean $\pm$ 95\% confidence interval.}
\renewcommand{\arraystretch}{1.3}
\setlength{\tabcolsep}{4pt}
\small

\begin{tabular}{l|c|cc|cc|cc|cc}
\toprule
\multicolumn{1}{c|}{\multirow{2}{*}{\shortstack{Map\\Type}}} 
& \multirow{2}{*}{Agents}
& \multicolumn{2}{c}{Success Rate (\%)} 
& \multicolumn{2}{c}{Makespan} 
& \multicolumn{2}{c}{Runtime (ms)} 
& \multicolumn{2}{c}{Reward} \\
\cline{3-4} \cline{5-6} \cline{7-8} \cline{9-10}
& & w/o & w/STEAM & w/o & w/STEAM & w/o & w/STEAM & w/o & w/STEAM \\
\midrule

\multirow{5}{*}{\shortstack{Dense\\maze}}
& 32  & \textbf{1.00±0.00} & \textbf{1.00±0.00} & 102.23±3.29 & \textbf{99.56±3.93}  & \textbf{97.69±2.64} & 112.74±1.38 & \textbf{247.31±12.10} & 233.35±11.24 \\
& 64  & 0.98±0.02 & \textbf{1.00±0.00} & 148.65±5.79 & \textbf{139.05±4.47} & \textbf{206.27±2.04} & 232.13±2.48 & \textbf{389.53±20.02} & 360.97±15.64 \\
& 128 & 0.53±0.09 & \textbf{0.75±0.08} & 232.72±5.23 & \textbf{219.35±5.58} & \textbf{413.69±3.21} & 476.70±43.20 & \textbf{555.29±23.79} & 542.41±20.20 \\
& 146 & 0.34±0.08 & \textbf{0.49±0.09} & 247.08±3.15 & \textbf{239.27±4.04} & \textbf{617.82±75.88} & 623.97±24.78 & \textbf{596.80±27.48} & 574.56±22.34 \\
& 160 & 0.19±0.07 & \textbf{0.34±0.08} & 250.99±2.29 & \textbf{246.09±3.20} & \textbf{536.58±10.96} & 586.06±24.44 & 585.17±27.41 & \textbf{574.13±23.02} \\

\midrule

\multirow{5}{*}{\shortstack{Ware\\house}}
& 128 & \textbf{1.00±0.00} & \textbf{1.00±0.00} & 141.00±3.61 & \textbf{128.63±2.57} & \textbf{317.29±1.59} & 378.87±2.25 & \textbf{365.90±15.10} & 320.79±10.54 \\
& 160 & \textbf{1.00±0.00} & \textbf{1.00±0.00} & 177.78±4.19 & \textbf{154.37±3.06} & \textbf{391.00±1.80} & 483.28±2.74 & \textbf{472.16±15.37} & 386.10±11.79 \\
& 172 & 0.98±0.02 & \textbf{1.00±0.00} & 191.30±4.38 & \textbf{168.50±3.58} & \textbf{425.17±2.00} & 525.41±3.15 & \textbf{505.45±16.13} & 423.75±13.48 \\
& 184 & 0.92±0.04 & \textbf{0.98±0.02} & 206.76±4.87 & \textbf{180.24±3.56} & \textbf{454.04±2.17} & 563.45±3.36 & \textbf{548.47±16.65} & 457.44±12.59 \\
& 192 & 0.86±0.06 & \textbf{0.99±0.01} & 213.67±4.54 & \textbf{186.80±3.82} & \textbf{467.21±2.01} & 556.97±9.04 & \textbf{561.11±14.42} & 472.95±13.74 \\

\bottomrule
\end{tabular}
\label{tab:primal-result}
\end{table*}
We integrate the proposed framework into each baseline and evaluate the resulting
congestion-aware variants under the same testing protocol. For all methods, we report
standard MAPF metrics, including \textbf{success rate}, \textbf{makespan}, and
\textbf{runtime}. For MAPF-GPT, we additionally report \textbf{sum of costs}; for PRIMAL2, we report the cumulative \textbf{reward}.
The metrics are defined as follows. 
The \textbf{success rate} is the proportion of episodes
in which all agents reach their goals within the maximum time limit. 
The
\textbf{makespan} is the maximum arrival time among all agents in an episode, averaged
over all test episodes. The \textbf{runtime} measures the average inference time per
timestep for all agents. The \textbf{sum of costs} is the total number of executed actions
until all agents reach their goals, and the \textbf{reward} is the cumulative
reinforcement-learning reward obtained by PRIMAL2 during an episode.

The detailed results are shown in Fig.~\ref{fig:result} and Table~\ref{tab:primal-result}.
Overall, the proposed method consistently improves the main MAPF performance metrics
across different baselines and environments. For MAPF-GPT, integrating our framework
substantially improves the success rate in dense random, maze, and warehouse settings,
while also reducing makespan and sum of costs in most cases. For PRIMAL2, as shown in
Table~\ref{tab:primal-result}, our method yields clear gains in success rate under both
dense maze and warehouse environments. The improvement becomes more pronounced as the
number of agents increases, indicating that the proposed congestion-aware guidance is
particularly beneficial in high-density scenarios where decentralized policies are more
likely to suffer from bottlenecks and local crowding.

\textbf{Generalization Across Baselines.}
The proposed framework exhibits strong generality across different learning-based
decentralized MAPF policies. It improves both MAPF-GPT and PRIMAL2, which differ in model
architecture, training procedure, and action-selection behavior. This supports the
policy-agnostic nature of our approach: 
inserted into its inference pipeline without retraining or architectural modification.
Rather than replacing the learned policy, our method provides structured test-time
guidance that preserves the original action preferences while reducing congestion.

\textbf{Out-of-Distribution Performance.}
We further observe that the proposed method is effective under distribution shift. In
particular, the warehouse environment is not included in the training distribution of
MAPF-GPT and therefore serves as an out-of-distribution setting. The baseline MAPF-GPT
models show degraded performance in this environment, especially under dense traffic.
After incorporating our congestion-aware enhancement, the success rate improves
substantially, and the MAPF-GPT-6M variant can reach nearly perfect success in the tested
warehouse setting. This result suggests that many failures under distribution shift are
related to congestion patterns such as aisle conflicts and bottleneck interactions, which
can be effectively mitigated by our spatial, temporal, and local correction mechanisms.

\textbf{Effect of Model Capacity.}
The improvements are not limited to small models. Both MAPF-GPT-2M and MAPF-GPT-6M benefit
from the proposed framework, indicating that the performance gain is not simply a
compensation for insufficient model capacity. Instead, our method introduces a
complementary mechanism that explicitly reasons about congestion at test time. This is
important because increasing model size alone does not necessarily resolve structured
coordination failures such as narrow-passage congestion or locally synchronized movements.

\textbf{Runtime and Reward Analysis.}
The proposed framework introduces only a small runtime overhead. This is expected because
the additional computation is limited to congestion detection, sparse cost-map correction,
and logit-level adjustment, without invoking centralized joint-state planning. In some
settings, the total runtime can remain close to that of the baseline because improved
coordination allows agents to reach their goals earlier, reducing the number of active
agents during later timesteps.

For PRIMAL2, we observe that the cumulative reward does not always increase, even when the
success rate and makespan improve. This is due to the reward design of PRIMAL2, where
waiting is penalized to discourage idling. Our method may intentionally encourage
short-term waiting or delayed movement to prevent future congestion. As a result, the
cumulative reward can slightly decrease, while the actual MAPF performance improves in
terms of success rate and makespan. This indicates a meaningful trade-off: sacrificing a
small amount of immediate reward can lead to better long-term coordination and higher
solution success.

\subsection{Agent Density Analysis}

To further examine whether the proposed method improves coordination by reducing local
crowding, we report an additional \emph{local agent density} metric during execution. This
metric measures how many other agents appear in the local observation window of each agent,
and therefore directly reflects the degree of local congestion experienced by decentralized
policies.  Formally, this metric is defined as:
\[
\rho = \frac{1}{|\mathcal{E}| \, T \, N}
\sum_{e \in \mathcal{E}} \sum_{t=1}^{T} \sum_{i=1}^{N}
\frac{1}{N-1} \sum_{j \neq i}
\mathbf{1}\big(\|p_i^t - p_j^t\| \le R\big),
\]
where $R$ denotes the observation radius, $T$ is the episode length,
$\mathcal{E}$ is the set of episodes, and $N$ is the number of agents.
The detailed results are reported in Table~\ref{tab:agent_density}. Across all tested
MAPF-GPT configurations, integrating our method consistently reduces the local density.
This reduction supports the main intuition behind our framework: the proposed spatial,
temporal, and local corrections do not merely improve success rate numerically, but also
change the execution behavior by dispersing agents away from crowded regions. The effect is
particularly visible in denser settings, such as Random-96 and Mazes-80, where congestion
is more likely to emerge. These results provide additional evidence that our method
proactively mitigates local aggregation, which helps explain the improvements in success
rate, makespan, and sum of costs reported above.

\begin{table}[t]
\centering
\caption{Agent density(\%) under different map types and model configurations. The numbers following each map type denote the corresponding agent count settings, while the first column specifies the MAPF-GPT model configuration used.}
\begin{tabular}{l|c|c|c|c}
\toprule
\textbf{MAPF-GPT} & \textbf{Random-64} & \textbf{Random-96} & \textbf{Mazes-64} & \textbf{Mazes-80} \\
\midrule
2M        & 0.22 & 0.35 & 0.24 & 0.31 \\ 
6M        & 0.22 & 0.34 & 0.21 & 0.28 \\ 
2M-STEAM  & 0.21 & \textbf{0.32} & 0.22 & 0.28 \\ 
6M-STEAM  & \textbf{0.20} & \textbf{0.32} & \textbf{0.20} & \textbf{0.26} \\
\bottomrule
\end{tabular}
\label{tab:agent_density}
\end{table}

\section{Conclusion}
This paper presented STEAM, a training-free congestion-aware enhancement framework for
learning-based decentralized MAPF. By incorporating spatial, temporal, and emergent local
congestion awareness, STEAM enables agents to anticipate and mitigate potential congestion
during test-time execution. Spatially resolvable congestion is handled by updating
agent-specific cost-to-go maps, while spatially unresolvable bottlenecks and emergent local
crowding are addressed through logit-level corrections.
STEAM is policy-agnostic and can be integrated into existing learning-based decentralized
MAPF algorithms without retraining or modifying their network architectures. Experiments
on multiple baselines and environments demonstrate consistent improvements in success rate
and coordination efficiency, especially in dense and congestion-prone settings, with only
minor additional computational overhead.

\addtolength{\textheight}{-12cm}   

\end{document}